\documentclass[runningheads]{llncs}
\usepackage[T1]{fontenc}
\usepackage{graphicx,verbatim,amsmath,multirow,amssymb,xcolor, array, placeins,orcidlink}
\begin{document}
\title{MMTA: Multi Membership Temporal Attention for Fine-Grained Stroke Rehabilitation Assessment}

\titlerunning{MMTA for Fine-Grained Stroke Rehabilitation Assessment}
%

\author{Halil Ismail Helvaci\inst{1} \and
Justin Huber\inst{2} \and
Jihye Bae \inst{1} \and
Sen-ching Samson Cheung \inst{1,3}}
\authorrunning{F. Author et al.}
%
\institute{Department of Electrical and Computer Engineering, University of Kentucky, Lexington, KY, USA \and
College of Medicine, University of Kentucky, Lexington, KY, USA \and
Department of Electrical and Computer Engineering, University of California, Davis, CA, USA \\
\email{halil.helvaci@uky.edu} }

\maketitle              
\begin{abstract}
To empower the iterative assessments involved during a person's rehabilitation, automated assessment of a person's abilities during daily activities requires temporally precise segmentation of fine-grained actions in therapy videos. Existing temporal action segmentation (TAS) models struggle to capture sub-second micro-movements while retaining exercise context, blurring rapid phase transitions and limiting reliable downstream assessment of motor recovery. We introduce Multi-Membership Temporal Attention (MMTA), a high-resolution temporal transformer for fine-grained rehabilitation assessment. Unlike standard temporal attention, which assigns each frame a single attention context per layer, MMTA lets each frame attend to multiple locally normalized temporal attention windows within the same layer. We fuse these concurrent temporal views via feature-space overlap resolution, preserving competing local contexts near transitions while enabling longer-range reasoning through layer-wise propagation. This increases boundary sensitivity without additional depth or multi-stage refinement. MMTA supports both video and wearable IMU inputs within a unified single-stage architecture, making it applicable to both clinical and home settings. MMTA consistently improves over the Global Attention transformer, boosting Edit Score by +1.3 (Video) and +1.6 (IMU) on StrokeRehab while further improving 50Salads by +3.3. Ablations confirm that performance gains stem from multi-membership temporal views rather than architectural complexity, offering a practical solution for resource-constrained rehabilitation assessment. 


\end{abstract}
\section{Introduction}

Stroke remains a leading cause of long-term disability in the United States, affecting over 795{,}000 individuals annually, with 77.4\% of survivors experiencing upper-limb impairment \cite{lawrence2001estimates,tsao2023heart}. Recovery of arm function is strongly tied to independence in activities of daily living, which makes accurate, repeatable measurement of motor performance central to rehabilitation. Yet routine clinical evaluation often relies on observation-based ordinal scales that are time-consuming, insensitive to subtle but clinically meaningful change, and only weakly correlated with real-world arm use \cite{murphy2011kinematic,waddell2017does,chen2021novel}. These limitations motivate automated approaches that transform long, continuous therapy recordings into clinically interpretable action units and high-resolution quantitative measures.

Temporal Action Segmentation (TAS) addresses this need by assigning an action label to each time step in untrimmed sequences, enabling automated assessment. Stroke rehabilitation presents a particularly challenging TAS regime: actions are fine-grained and visually subtle, and clinically meaningful transitions may occur within only a few frames at sub-second timescales \cite{kaku2022strokerehab}. Global self-attention can capture long-range temporal structure, but as temporal context grows, softmax normalization disperses attention across all frames, diluting local boundary evidence, an inherent limitation we term the \emph{temporal granularity bottleneck}. Prior work mitigates boundary errors using multi-stage refinement \cite{farha2019ms,ishikawa2021alleviating}, hierarchical temporal encoders \cite{yi2021asformer}, or sparsity and locality constraints in attention \cite{bahrami2023much,beltagy2020longformer,gulati2020conformer,liu2021swin,lu2024fact,van2023aspnet,zhang2022actionformer}. However, these approaches produce one normalized update per frame per layer, forcing a sub-optimal resolution of competing context around transitions \cite{farha2019ms,helvaci2024localizing,yi2021asformer,bahrami2023much}. We argue that preserving multi-membership context is critical for fine-grained TAS: near action boundaries, frames often contain ambiguous evidence spanning multiple phases; standard attention collapses this into a single summary, whereas in reality, a singe frame could simultaneously preserves multiple local contextual views that would be useful for robust segmentation.

In this work, we introduce Multi-Membership Temporal Attention (MMTA), a boundary-preserving attention operator for fine-grained TAS that eliminates the need for multi-stage refinement and global attention. Unlike standard windowed attention, which yields a single locally normalized output per frame, MMTA allows each frame to participate in $N$ overlapping local windows per layer, yielding multiple locally normalized, window-conditioned updates that are fused via an explicit overlap-resolution rule. This preserves competing contextual evidence near transitions and improves boundary localization. We evaluate on Stroke Rehabilitation therapy recordings (StrokRehab video and IMU) \cite{kaku2022strokerehab} and 50Salads \cite{stein2013combining}, where MMTA consistently outperforms existing TAS methods on boundary-sensitive metrics including Edit Score and Action Error Rate.

\section{Method}
\label{sec:method}

\textbf{Problem Setup.} Given an untrimmed sequence (video or wearable sensors) represented as frame-level feature vectors
$\mathbf{X}=\{\mathbf{x}_t\}_{t=1}^{T}$, TAS predicts a label for each time step
$\mathbf{Y}=\{y_t\}_{t=1}^{T}$ over $C$ action classes.
An action instance is a tuple $y_i=(s_i,e_i,a_i)$ with onset $s_i$, offset $e_i$, and class label $a_i\in\{1,\dots,C\}$, where $1\le s_i < e_i \le T$.
Accurate TAS therefore requires both correct labeling and precise boundary localization.

\begin{figure}[t]
    \centering
    \includegraphics[width=0.9\linewidth]{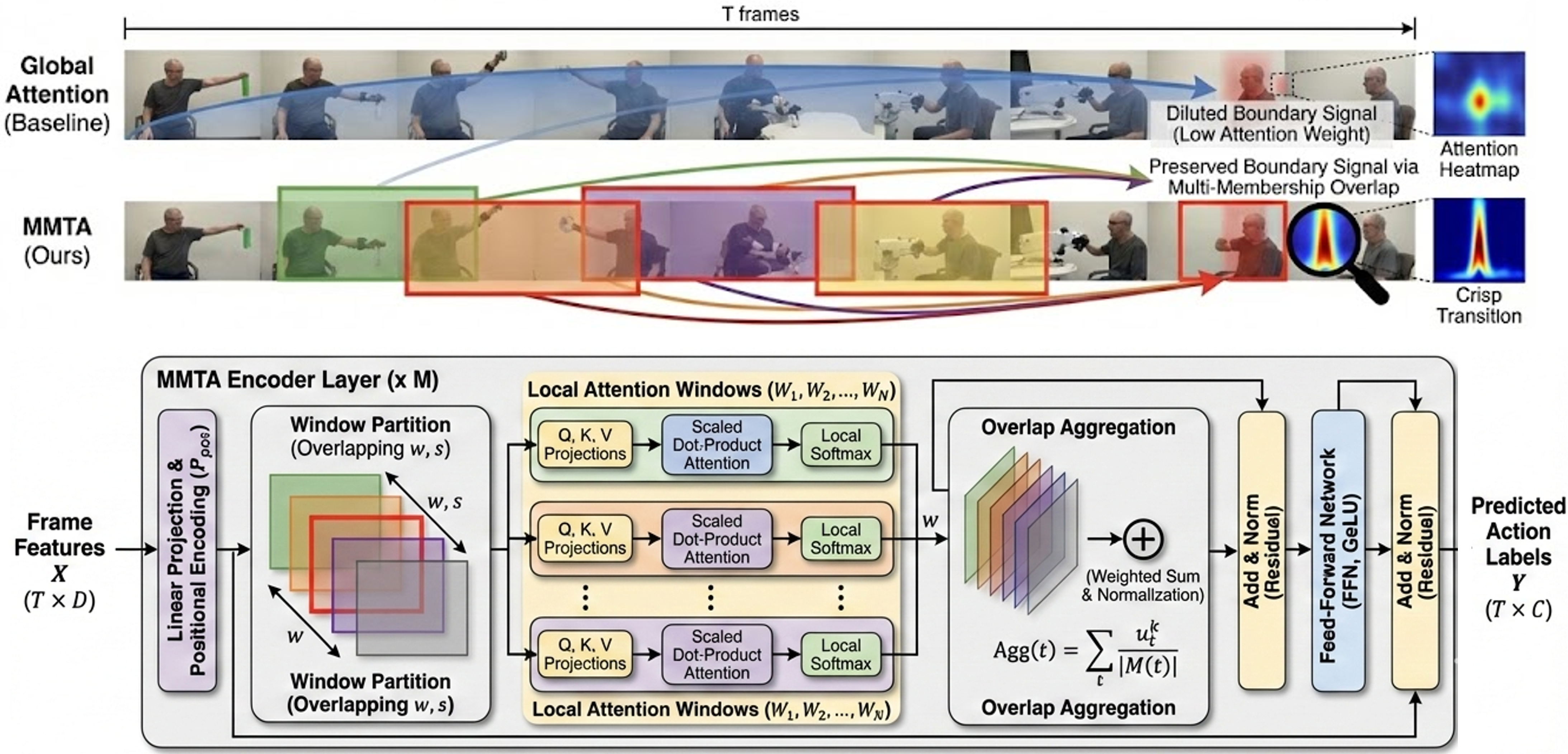}
    \label{fig:model}
    \caption{MMTA replaces global attention with locally normalized attention over overlapping temporal windows. Frames in overlap regions receive multiple window-conditioned updates, reconciled via overlap-resolution aggregation.}
\label{fig:model}
\end{figure}

\noindent\textbf{Overview.} We adopt a single-stage transformer encoder backbone for TAS and replace global self-attention with MMTA, as illustrated in Fig.~\ref{fig:model}. MMTA assigns each frame to multiple local overlapping windows, produces locally normalized updates near boundaries, and reconciles the frames with explicit overlap resolution, allowing context to propagate across layers without global $T\times T$ attention.

\noindent\textbf{The Temporal Granularity Bottleneck.} Global self-attention computes attention across the entire temporal sequence $T$. While effective for capturing coarse activities, this global normalization disperses attention more evenly across all frames. For a boundary frame $t_b$ marking an action transition, the attention weight from query frame $t_b$ to a nearby key frame $t_b \pm \delta$ (where $\delta$ is small, e.g., 2-5 frames) is:

\begin{equation}
Att(t_b,t_b\pm\delta)=
\frac{\exp(s_{t_b,t_b\pm\delta})}{\sum_{j=1}^{T}\exp(s_{t_b,j})},\qquad
s_{i,j}=\frac{\mathbf{q}_i^\top\mathbf{k}_j}{\sqrt{d_k}} ,
\end{equation}
where $\mathbf{q}_t,\mathbf{k}_j,\mathbf{v}_j\in\mathbb{R}^{d_h}$ are query, key, and value vectors in a single head, and $s_{i,j}$ denotes the scaled dot-product. In fine-grained TAS, boundary evidence is often brief and localized. As $T$ grows, the softmax denominator increases and the probability mass assigned to any fixed local neighborhood shrinks, even when local similarity is high. 

Formally, the expected local attention weight decays as $\mathbb{E}[A(t_b, t_b \pm \delta)] = O(1/T)$ under standard softmax attention, reducing sensitivity to boundary transitions. This attention dilution reduces sensitivity to short transitions and leading to temporal over-smoothing. Because the effect is driven by normalization over $T$ elements, scaling model depth or width does not directly recover boundary fidelity. We refer to this limitation as the \emph{temporal granularity bottleneck}.

\noindent\textbf{Multi-Membership Temporal Attention (MMTA).} Unlike standard windowed attention, which produces a single locally normalized representation per frame, MMTA generates multiple independently normalized representations per frame within each layer and reconciles them explicitly. It mitigates dilution by restricting normalization to local windows while preserving cross-window continuity via overlap. To model local temporal dependencies while maintaining contextual continuity across segments, we restrict the self-attention operation to partially overlapping temporal windows. Given an input sequence of length $T$, we partition it into $N = \lceil T / (w - o) \rceil$ windows, each covering $w$ frames with an overlap of $o$ frames between consecutive windows with stride $s = w - o$. Let $\mathcal{W}_i$ denote the index set of frames in the $i^{th}$ window. Due to overlap, a frame $t$ may belong to multiple windows; its membership set is $\mathcal{M}(t)=\{i \mid t\in\mathcal{W}_i\}$ and $m(t)=|\mathcal{M}(t)|\ge 1$ is the size of membership set. For window $i$, the query, key, and value projections are denoted as $Q_i, K_i, V_i \in \mathbb{R}^{(w) \times d}$.

By limiting normalization to local neighborhoods of size $w \ll T$, MMTA replaces the global denominator $\sum_{j=1}^{T}\exp(s_{i,j})$ with a localized one $\sum_{j=1}^{w}\exp(s_{i,j})$, preventing the dilution of local similarities and restoring sharp boundary attention around transitions. The attention operation \cite{vaswani2017attention} within each window follows a scaled dot-product attention as follows: $\text{Attn}(Q_i, K_i, V_i) = \text{Softmax}\left( \frac{Q_i K_i^\top}{\sqrt{d_k}} \right) V_i
\label{eq:attn}$, where $d_k$ is the key dimensionality used for scaling and $i = 1, \dots, N$, is the index to each overlapping wondow. This formulation ensures that attention is computed only among frames within each overlapping local region, allowing information exchange across adjacent windows via the shared overlap.

Restricting attention to a single window may introduce boundary effects at window edges. MMTA leverages overlap to obtain multiple locally normalized views of the same frame within a layer: if $t$ belongs to $m(t)$ windows, it produces $m(t)$ window-conditioned updates. MMTA then applies overlap-resolution fusion to reconcile these updates into one representation per frame, facilitating cross-window context transfer without the diluting effect of global attention.

\noindent\textbf{Overlap Resolution Aggregation.} Windowed attention produces an output for each position inside $\mathcal{W}_i$. If a frame $t$ appears in multiple windows, MMTA yields multiple window-conditioned outputs $\{\mathbf{u}_t^{(i)}\}_{i\in\mathcal{M}(t)}$. We reconcile the multi-membership with an explicit overlap-resolution operator.
\begin{equation}
\tilde{\mathbf{h}}_t = \text{Agg}\!\left(\{\mathbf{u}_t^{(i)}\}_{i\in\mathcal{M}(t)}\right) = \frac{1}{m(t)}\sum_{i\in\mathcal{M}(t)}\mathbf{u}_t^{(i)}.
\label{eq:overlap_agg}
\end{equation}
Unlike standard windowed attention, which yields one normalized update per frame per layer, MMTA constructs multiple locally normalized updates and resolves them explicitly. Although attention is local within each window, overlap allows information to pass between neighboring windows. Stacking $M$ MMTA layers expands the effective receptive field as $w+(M-1)s$ frames, providing longer-range context without global $T\times T$ attention.

\noindent\textbf{MMTA Operator.} Let $\mathbf{H}\in\mathbb{R}^{T\times d}$ be the input sequence and $\{\mathcal{W}_k\}_{k=1}^{N}$ the overlapping temporal windows. For each window $k$, we compute window-restricted attention $U_k = \text{Attn}(\mathbf{H}{\mathcal{W}_k})$, and reconcile multi-membership frames via Eq.~\ref{eq:overlap_agg} to yield the MMTA operator as:
\begin{equation}
        \text{MMTA}(\mathbf{H}) \;:=\; \left\{\frac{1}{m(t)}\sum_{k\in\mathcal{M}(t)}\mathbf{u}_t^{(k)}\right\}_{t=1}^{T}
\in\mathbb{R}^{T\times d}. 
\end{equation} 
This aggregation produces a single representation per frame while enabling information transfer across neighboring windows through shared overlap.

\noindent\textbf{MMTA Encoder Layer.} At layer $\ell$, we apply:
\begin{equation}
\tilde{\mathbf{H}}^{(\ell)} = \text{MMTA}\!\left(\mathbf{H}^{(\ell-1)}\right),
\end{equation}
\begin{equation}
\mathbf{H}^{(\ell)} = \tilde{\mathbf{H}}^{(\ell)} + \text{FFN}\!\left(\text{LayerNorm}(\tilde{\mathbf{H}}^{(\ell)})\right).
\end{equation}

\noindent\textbf{Complexity.} Global self-attention costs $O(T^2d)$ operations per layer. MMTA attends within $N\approx T/s$ windows, each of cost $O(w^2d)$, giving $O\!\left(\frac{T}{s}w^2d\right)$. For fixed $(w,s)$, MMTA scales linearly with sequence length $T$, in contrast to the quadratic scaling of global self-attention.

\section{Experiments}
\label{sec:experiments}

\textbf{Datasets.}
We evaluate MMTA on StrokeRehab \cite{kaku2022strokerehab} (Video and IMU modalities), a clinically grounded upper-limb therapy dataset where sub-second accuracy in transition timing is critical for assessing motor recovery, and on the public 50Salads benchmark \cite{stein2013combining} to test generalization beyond stroke rehabilitation. \textbf{StrokeRehab (Video/IMU)} contains 3,372 trials from 51 stroke-impaired patients and 20 healthy subjects, with 120,891 annotated functional primitives spanning nine daily activities (e.g., feeding, brushing teeth). Labels were produced by trained coders under expert supervision, with a a high inter-rater reliability (Cohen’s $\kappa\ge0.96$). IMU data comprise 76 kinematic channels from nine sensors sampled at 100\,Hz; video was recorded from two orthogonal cameras (1088$\times$704, 60/100 fps). Deep-learned I3D features \cite{carreira2017quo} of the videos are used in our study as the raw videos were not released for privacy reasons. The dataset is accessible on SimTK: \textit{\url{https://simtk.org/projects/primseq}}. \textbf{50Salads} includes 50 salad-preparation videos (downsampled to 15fps) with 17 action classes (e.g., cut tomato, add salt), performed by 25 subjects. This dataset provides a standard benchmark for temporal action segmentation in non-clinical settings. I3D features extracted from the videos are used as inputs to our network.

\begin{table}[t]
\caption{Comparison with state-of-the-art methods on StrokeRehab (top) and 50Salads (bottom). Both transformer baselines share an identical backbone; \textit{Global Attention} uses full self-attention, while \textit{MMTA (ours)} replaces it with the proposed module.}
\centering
\scriptsize

\resizebox{0.6\columnwidth}{!}{
\begin{tabular}{l|l|c|c|c|c}
\hline
\multirow{2}{*}{Backbone} & \multirow{2}{*}{Model} & \multicolumn{2}{c|}{\textbf{Video}} & \multicolumn{2}{c}{\textbf{IMU}} \\ \cline{3-6}
 &  & ES $\uparrow$  & AER $\downarrow$ & ES $\uparrow$  & AER $\downarrow$  \\
\hline
\multirow{3}{*}{TCN} 
 & MS-TCN* \cite{farha2019ms}  & 60.7 & 0.408 & 66.9 & 0.372 \\
 & MS-TCN \cite{farha2019ms}   & 62.2 & 0.392 & 68.9 & 0.330 \\
 & MS-TCN$^{+}$ \cite{farha2019ms} & 62.7 & 0.390 & 68.8 & 0.317 \\ 
\hline
\multirow{2}{*}{Boundary} 
 & ASRF* \cite{ishikawa2021alleviating}  & 56.9 & 0.449 & 68.2 & 0.328 \\
 & ASRF \cite{ishikawa2021alleviating}   & 58.7 & 0.436 & 67.9 & 0.349 \\
\hline
\multirow{2}{*}{Seq2Seq} 
 & Seg2Seq \cite{kaku2022strokerehab}    & 67.6 & 0.322 & 63.0 & 0.337 \\
 & Raw2Seq \cite{kaku2022strokerehab}    & 66.6 & 0.329 & 68.8 & 0.305 \\ 
\hline
\multirow{2}{*}{Transformer} 
 & Transformer (Global)                & 69.8 & 0.302 & 68.9 & 0.311 \\ 
 & MMTA (ours)          & \textbf{71.1} & \textbf{0.289} & \textbf{70.5} & \textbf{0.295} \\ 
 \hline
\end{tabular}}
\label{tab:stroke_rehab}

\vspace{0.6em}

\resizebox{0.5\columnwidth}{!}{
\begin{tabular}{l|l|c|c}
\hline
\multirow{2}{*}{Backbone} & \multirow{2}{*}{Model} 
  & \multicolumn{2}{c}{\textbf{50Salads}} \\ \cline{3-4}
 &  & ES $\uparrow$  & AER $\downarrow$ \\
\hline
\multirow{3}{*}{TCN}
 & MS-TCN* \cite{farha2019ms}      & 68.8 & 0.47 \\
 & MS-TCN \cite{farha2019ms}       & 70.8 & 0.43 \\
 & MS-TCN$^{+}$ \cite{farha2019ms} & 76.4 & 0.32 \\ 
\hline
\multirow{2}{*}{Boundary}
 & ASRF* \cite{ishikawa2021alleviating}  & 74.0 & 0.34 \\
 & ASRF \cite{ishikawa2021alleviating}   & 75.2 & 0.33 \\
\hline
\multirow{2}{*}{Seq2Seq}
 & Seg2Seq \cite{kaku2022strokerehab}    & 76.9 & 0.30 \\
 & Raw2Seq \cite{kaku2022strokerehab}    & 69.4 & 0.54 \\ 
\hline
\multirow{2}{*}{Diffusion} 
 & DiffAct \cite{liu2023diffusion}       & 85.0 & - \\
 & DiffAct++ \cite{diffact++}     & 85.8 & - \\
\hline
\multirow{7}{*}{Transformer}
 & ASFormer \cite{yi2021asformer}        & 79.6 & - \\
 & ASPnet \cite{van2023aspnet}           & 87.5 & - \\
 & BaFormer \cite{wang2024efficient}     & 84.2 & - \\
 & LTContext \cite{bahrami2023much}      & 83.2 & - \\
 & Transformer (Global)                  & 85.1 & 0.149 \\ 
 & \textbf{MMTA (ours)}                  & \textbf{88.4} & \textbf{0.116} \\ 
\hline
\end{tabular}}
\label{tab:50salads}

\end{table}

\noindent\textbf{Implementation Details.} We follow the same splits and standard protocols from prior work \cite{farha2019ms,ishikawa2021alleviating,kaku2022strokerehab,yi2021asformer}. Video features are extracted using I3D \cite{carreira2017quo}. In order to reflect differences in temporal resolution and action duration across modalities, we use dataset-specific window length and stride \((w,s)\) with \((200,10)\) for StrokeRehab Video, \((500,10)\) for StrokeRehab IMU, and \((1500,500)\) for 50Salads. Model capacity is tuned per dataset. StrokeRehab uses 3 encoder layers with 4 heads and hidden size 512. 50Salads uses 3 layers with 2 heads and hidden size 256.

StrokeRehab is trained for 25 epochs with batch size 2 and initial learning rate \(10^{-3}\), which is reduced by a factor of 0.01 if the validation focal loss plateaus for 5 epochs. 50Salads is trained for 10 epochs with batch size 2. We use SGD with momentum 0.9 and weight decay \(10^{-4}\), dropout 0.2, focal loss with \(\alpha{=}0.25\) and \(\gamma{=}2\), and gradient clipping with max-norm 5 for StrokeRehab and 6 for 50Salads. All experiments are run on an NVIDIA RTX A6000. No multi-stage refinement or post-processing is used unless explicitly noted.

\noindent\textbf{Evaluation Metrics.} We report the segmentation Edit Score (ES) and Action Error Rate (AER), both computed from the Levenshtein distance \(L\) between the ground-truth segment transcript \(G\) and the predicted segment transcript \(P\) \cite{farha2019ms,yi2021asformer,ishikawa2021alleviating,kaku2022strokerehab}. For StrokeRehab, we additionally report per-class sensitivity, specificity, and F1 to provide complementary insight on segmenting clinical meaningful action primitives.


\noindent\textbf{Comparison with State-of-the-Art Methods.} Table~\ref{tab:stroke_rehab} compares MMTA against representative baselines across model families, all using a single-stage transformer encoder. Baselines span five paradigms: TCNs with temporal 1D convolutions (MS-TCN~\cite{farha2019ms}); boundary-aware models that explicitly predict action boundaries (ASRF~\cite{ishikawa2021alleviating}); Seq2Seq methods that auto-regressively generate segment/label sequences (Seg2Seq, Raw2Seq~\cite{kaku2022strokerehab}); Transformers for long-range temporal modeling (ASFormer~\cite{yi2021asformer}, ASPnet~\cite{van2023aspnet}, BaFormer~\cite{wang2024efficient}, and variants~\cite{beltagy2020longformer,bahrami2023much}); and diffusion-based methods that cast segmentation as iterative denoising (DiffAct~\cite{liu2023diffusion}). Transformer (Global) serves as our direct baseline, sharing the identical architecture as MMTA but replacing multi-membership temporal attention with full self-attention. Output logits are optionally smoothed before $argmax$ to reduce frame noise ($+$: smoothing applied; $*$: selected by best validation frame accuracy).

On StrokeRehab (video), performance improves consistently from boundary-aware methods to TCN, Seq2Seq, and transformers, reflecting stronger long-range temporal modeling for subtle transitions. IMU results are more clustered, suggesting local kinematic cues reduce reliance on long-range context, though transformers still lead. Across both modalities, MMTA consistently outperforms global attention by 1.3–1.6 ES and lowers AER, indicating better boundary localization. On 50Salads, MMTA is best among methods reporting both metrics, improving over Transformer (Global) and surpassing prior ES and AER results (e.g., ASPnet, DiffAct++). 

\begin{figure}[t]
    \centering
    \includegraphics[width= 0.7\linewidth]{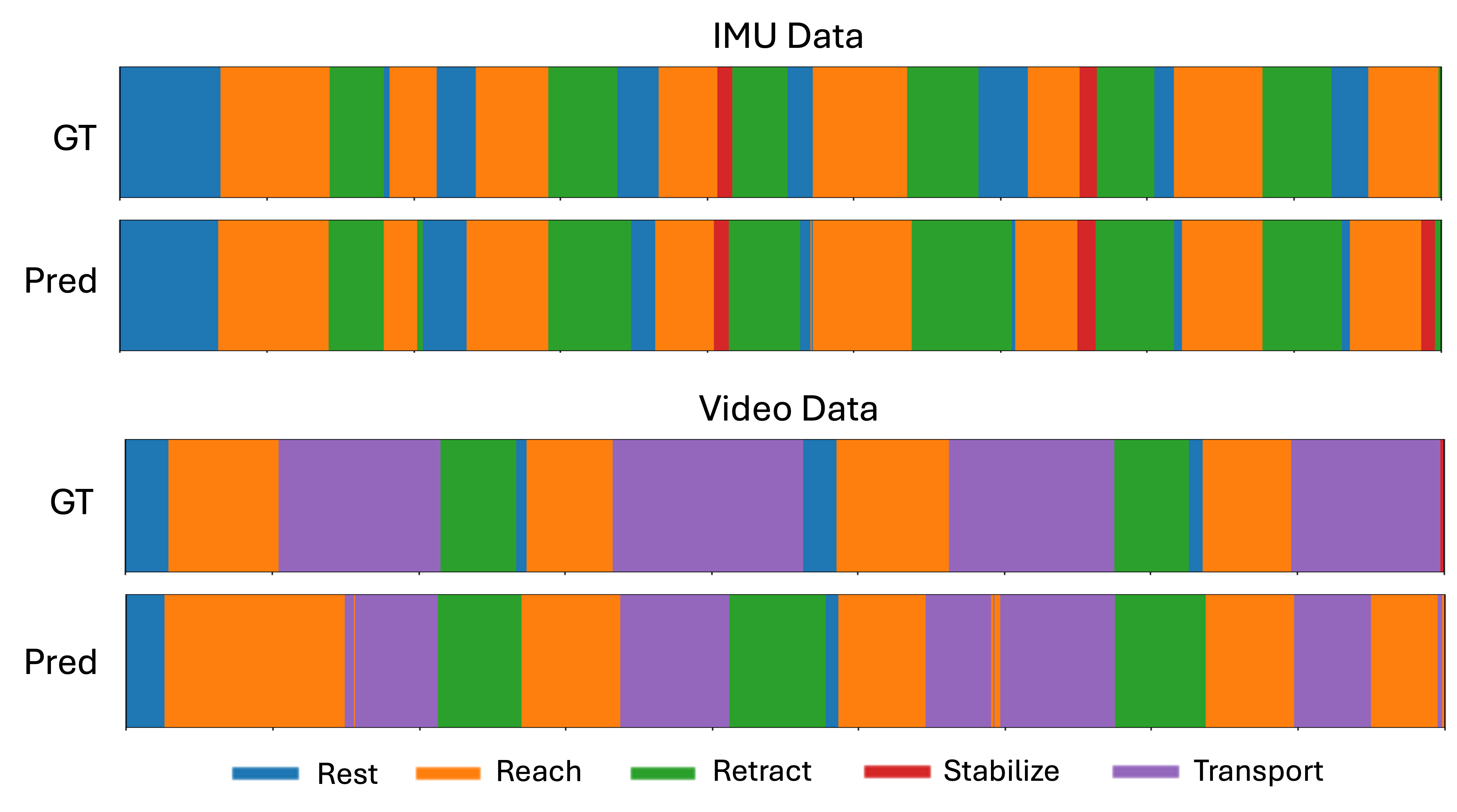}
    \caption{Qualitative comparison of predicted (Pred) and ground-truth (GT) action labels on an example test sequence for IMU data (top) and video data (bottom).}
    \label{fig:combined_predictions}
\end{figure}

Fig.~\ref{fig:combined_predictions} visualizes predictions versus ground truth. MMTA produces more accurate boundary transitions with fewer spurious segments, while preserving the overall temporal structure of action sequences. Errors concentrate in visually or kinematically ambiguous primitives such as the rest phase between action.


\noindent\textbf{Efficiency.} We measure GPU memory usage on 50Salads and find that MMTA requires only 422–460 MB across window sizes, compared to 1.7 GB for MS-TCN and 3.5 GB for ASFormer. These techniques require substantially more memory because they rely on either multi-stage processing (MS-TCN) or global self-attention over the full sequence (ASFormer).

\begin{table}[!ht]
\caption{Evaluation of frame-level action classification across the StrokeRehab Video and IMU datasets}
\centering
\scriptsize
\resizebox{0.6\columnwidth}{!}{
    \begin{tabular}{l|c c c|c c c}
    \hline
         \multirow{2}{*}{\textbf{Action}} & \multicolumn{3}{c|}{\textbf{Video}} & \multicolumn{3}{c}{\textbf{IMU}} \\ \cline{2-7}
              & Sens. & Spec. & F1 & Sens. & Spec. & F1 \\ \hline
    Rest      & 0.70 & 0.92 & 0.66 & 0.69 & 0.93 & 0.67 \\ 
    Reach     & 0.52 & 0.96 & 0.59 & 0.59 & 0.94 & 0.64 \\ 
    Retract   & 0.60 & 0.97 & 0.62 & 0.64 & 0.97 & 0.69 \\ 
    Stabilize & 0.59 & 0.91 & 0.63 & 0.65 & 0.90 & 0.60 \\ 
    Transport & 0.77 & 0.85 & 0.70 & 0.72 & 0.88 & 0.71 \\ \hline
    \textbf{Macro Avg.} 
              & 0.64 & 0.92 & 0.64
              & 0.66 & 0.92 & 0.66 \\ \hline
    \end{tabular}
}
\label{tab:cls_combined}
\end{table}



\noindent\textbf{Classification Performance.} Beyond segmentation quality, we evaluate frame-level classification performance to assess modality-specific strengths across all datasets. Table~\ref{tab:cls_combined} reports sensitivity, specificity, and F1-scores for key primitives on StrokeRehab. IMU signals generally yield higher sensitivity and F1 for dynamic primitives such as \textit{Reach} and \textit{Retract}, reflecting their advantage in capturing fine-grained motion dynamics at high sampling rates. In contrast, the video modality consistently achieves higher specificity, reducing false positives, particularly for visually distinctive actions like \textit{Transport}.

\noindent\textbf{Ablation Study.} Table~\ref{table:window_ablation} evaluates the impact of window size and stride on StrokeRehab. Smaller strides consistently improve boundary localization due to higher number of multi-membership frames near transitions, while optimal window size differs by modality: video peaks at w=200 and IMU at w=500, reflecting differences in temporal resolution and motion dynamics. Performance degrades at both extremes, confirming that modality-specific window tuning is important for balancing local sensitivity and contextual coverage.


\begin{table}[!ht]
\caption{Ablation study on the impact of window size and stride on the StrokeRehab Video and IMU datasets.}
\centering
\scriptsize
\resizebox{0.6\columnwidth}{!}{
\begin{tabular}{l|c|c|c|c|c|c}
\hline
\multicolumn{7}{c}{StrokeRehab} \\ \hline
\multirow{2}{*}{Model} & \multicolumn{2}{c|}{Video} & \multicolumn{2}{c|}{IMU} & \multirow{2}{*}{Window Size} & \multirow{2}{*}{Stride} \\ \cline{2-5} & ES $\uparrow$ & AER $\downarrow$ & ES $\uparrow$ & AER $\downarrow$ & & \\ \hline
\multirow{6}{*}{MMTA} 
 & 62.0 & 0.388 & 67.6 & 0.324 & 1500 & 500 \\
 & 66.8 & 0.332 & 68.9 & 0.314 & 1000 & 500 \\
 & 68.2 & 0.318 & 66.9 & 0.331 & 800  & 500 \\
 & 68.7 & 0.313 & \textbf{70.5} & \textbf{0.295} & 500  & 10 \\
 & \textbf{71.1} & \textbf{0.289} & 66.8 & 0.332 & 200  & 10 \\
 & 68.0 & 0.320 & 63.8 & 0.364 & 100  & 10 \\
\hline
\end{tabular}}
\label{table:window_ablation}
\end{table}

\section{Conclusion}
We proposed MMTA, a multi-membership temporal attention operator that addresses the temporal granularity bottleneck by allowing each frame to participate in multiple overlapping local windows, preserving competing boundary evidence without multi-stage refinement or global attention, MMTA consistently improves Edit Score and Action Error Rate over strong baselines with reduced complexity. Its linear complexity and low memory footprint make MMTA a practical solution for automated rehabilitation assessment in clinical and home settings. A limitation of MMTA is the reliance on fixed window configurations, which may not optimally adapt to varying temporal dynamics; future work will explore adaptive or learned windowing strategies.

\begin{credits}
\subsubsection{\ackname} Research reported in this publication was supported by the Igniting Research Collaborations (IRC) at the University of Kentucky.
\subsubsection{\discintname}
The authors have no competing interests to declare that are relevant to the content of this article.
\end{credits}

%
%
%
\bibliographystyle{splncs04}
\bibliography{main}

\end{document}